\documentclass[runningheads]{llncs}
\usepackage[T1]{fontenc}
\usepackage{graphicx}

\usepackage[hidelinks,breaklinks=true,bookmarks=false]{hyperref}
\usepackage{color}

\usepackage{tabularx}
\usepackage{bbding}
\begin{document}
%
\newif\ifanonymous
\anonymousfalse
\newif\ifshowresultsdiscussion
\showresultsdiscussiontrue

\title{\ifanonymous {\ifshowresultsdiscussion \large [Exp] \fi Contrastive Learning for Authorship Verification}\else Contrastive Learning for Authorship Verification\fi}

\ifanonymous

  \author{\phantom{First Author\orcidID{0000-1111-2222-3333}\Envelope}}
  \authorrunning{\phantom{First Author}}

  \institute{
    \phantom{Anon University}\\
    \phantom{anon@anon.edu}
  }

\else
  \author{Peter Kirby\orcidID{0009-0003-8172-8932}\Envelope}
  \authorrunning{Peter Kirby}

  \institute{
    Georgia Institute of Technology \\
    \email{pkirby6@gatech.edu}
  }

\fi

\maketitle
\begin{abstract}
\ifshowresultsdiscussion
   Our results show that contrastive learning outperforms a classification-based approach to authorship verification under the tested settings. We identify loss function, batch size, training duration, pre-trained model, input context length, and random text span data augmentation as important factors of model performance. Based on these considerations, we develop a ModernBERT Bi-Encoder model that achieves 98.4\% accuracy on the PAN21 authorship verification task.
\else
   This paper studies the choices involved in fine-tuning transformer models for authorship verification. We compare contrastive-based and classification-based training approaches under controlled conditions. We explore the performance impact of loss function, input context length, batch size, pre-trained model, training duration, and data augmentation with PAN21 fanfiction data. An authorship verification model based on these considerations is trained for comparison to prior work.
\fi

\end{abstract}
%
\section{Introduction}
How should authorship verification models be trained? As a task that asks whether two texts are by the same author~\cite{koppel2004authorship}, it can easily be approached as a binary classification problem. When used for classification, transformer models can be trained as cross-encoders that use attention between two texts, while contrastive models such as bi-encoders~\cite{humeau2020polyencoders} learn representations by training on many pairwise comparisons efficiently. Both approaches have a plausible story for why they might perform better than the other.

PAN20~\cite{kestemont2020overview} provided a large authorship verification dataset based on fanfiction stories, and PAN21~\cite{kestemont2021pan21av} involved an open-set problem of testing on unseen authors and unseen fandoms. This setting continues to appear in work on datasets for evaluation of authorship verification systems~\cite{brad2022rethinking,sawatphol2024addressing,tyo2023valla}, and studies continue to report results on PAN21~\cite{li2025tame,mirallesgonzalez2025llm,nguyen-etal-2023-improving}. It is accordingly useful for a comparison to prior work on the authorship verification task.

While transformer models are effective at authorship verification, it is less clear which fine-tuning choices are responsible for their performance. In the PAN21 fanfiction setting, Tyo et al.~\cite{tyo2021siamese} fine-tuned BERT with pairwise contrastive loss, while Peng et al.~\cite{peng2021encoding} showed improved performance with a classification approach that averaged BERT representations from paired snippets to work around BERT's limited context length. Nguyen et al.~\cite{nguyen-etal-2023-improving} later achieved competitive PAN21 results with binary classification using BigBird~\cite{zaheer2020bigbird}, a longer-context transformer. Yet the top overall PAN21 system from Boenninghoff et al.~\cite{boenninghoff2021o2d2,kestemont2021pan21av} had a contrastive loss objective and used a bidirectional LSTM with attention~\cite{boenninghoff2019explainable}. Prior results point to the importance of input context length for PAN21. It still remains unclear whether binary classification or contrastive learning would be more effective when other factors are controlled, such as the pretrained model and input context length.

Li et al.~\cite{li2025tame} clarify several design choices in this context, including the importance of fine-tuning, the benefit of cased tokenization, and the effectiveness of cosine distance when applied after mean pooling. However, the choice of training objective was explicitly left to future work, including whether contrastive learning shows improvement over a standard classification objective.

Our main contribution is a comparison of contrastive and classification-based training for transformer-based authorship verification. We tune batch size and learning rate separately for each approach, while holding the remaining training conditions fixed to support a meaningful comparison. We perform this comparison across several model architectures, as shown in Table~\ref{tab:model_comparison}.

\begin{table}
\caption{Transformer Model Size Summary}
\label{tab:model_comparison}
\centering
\scriptsize
\setlength{\tabcolsep}{3pt}
\begin{tabular}{|p{4.0cm}|p{1.7cm}|p{1.7cm}|p{1.7cm}|}
\hline
\textbf{Model} & \textbf{Parameters} & \textbf{Layers} & \textbf{Dimensions} \\
\hline
TinyBERT~\cite{jiao-etal-2020-tinybert} & 15M & 4 & 312 \\
\hline
DistilBERT-cased~\cite{sanh2020distilbertdistilledversionbert} & 66M & 6 & 768 \\
\hline
BERT-base-cased~\cite{devlin2019bert} & 110M & 12 & 768 \\
\hline
ModernBERT-base~\cite{warner2024smarterbetterfasterlonger} & 149M & 22 & 768 \\
\hline
ModernBERT-large~\cite{warner2024smarterbetterfasterlonger} & 395M & 28 & 1024 \\
\hline
\end{tabular}
\end{table}

We also identify some of the practical choices that matter on this task. Various contrastive loss functions are considered. We further explore the impact of data augmentation, model choice, training duration, and input context length in the PAN21 authorship verification setting. We release the code and the dataset in Parquet format, including the validation data split used, at \url{https://github.com/petekirby/contrastive-av} to support future research.


\section{Methodology}

\subsection {Data Augmentation}
``The best way to make a machine learning model generalize better is to train it on more data.''~\cite[p.~240]{goodfellow2016deep} Following Boenninghoff et al.~\cite{boenninghoff2021o2d2}, we convert the training data from fixed pairs into individual document rows with author IDs for dynamic pair recombination, yielding varied same-author and different-author pairs, with billions of potential distinct negative pairs. We use PyTorch Metric Learning~\cite{musgrave2020pytorchmetric} to sample authors randomly, each time randomly selecting two document samples per author. Each document sample is rotated to start at a random word boundary before truncation by the tokenizer. The goal was to retain fine details for this task, but a recent study~\cite{gonzalezmarquez2026cropping} suggests a general advantage for text cropping over SimCSE-style dropout~\cite{gao-2021-simcse}. Random text rotation as a data augmentation technique allows equal length truncated spans at any position with little disturbance to the data. This can create thousands of distinct views for one sample and millions of distinct pairs of strings for a pair of samples.
\begin{figure}[t]
    \centering
    \includegraphics[width=\textwidth]{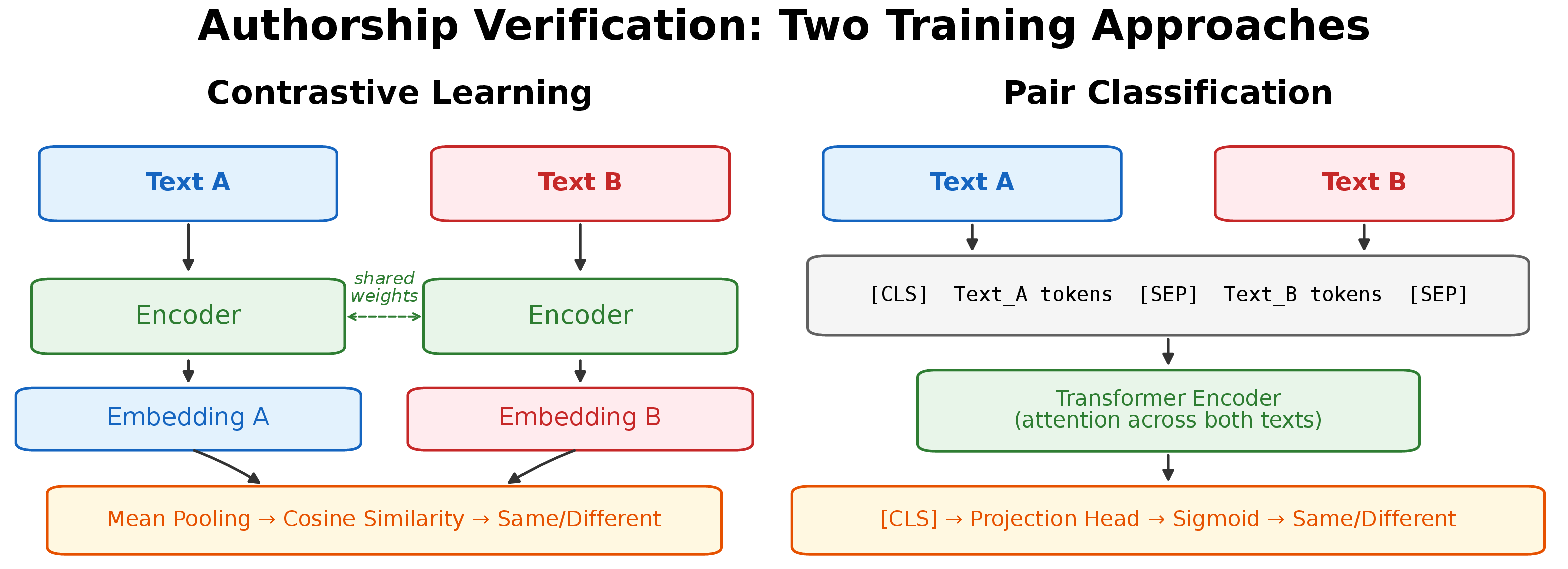}
    \caption{Comparison of our two training approaches. Contrastive learning embeds each text independently and compares the resulting embeddings by cosine similarity. Pair classification concatenates both texts into one input sequence, allowing the transformer encoder to use attention across the two texts.}
    \label{fig:comparison-approach}
\end{figure}


\subsection{Pair Classification}
We fine-tune a transformer model for sequence classification, including weights for the projection head included with the model. The cross-encoder concatenates two truncated tokenized texts as a single input. Each document has one positive pair and a negative, different-author pair selected from the batch at random. A learning rate multiple of 5 is used for the projection head~\cite{chernyavskiy2021transformers,joshi2019bertcoref}. For prediction, the sigmoid output threshold that maximizes F1 on the validation set is used.

\subsection {Contrastive Learning}
As shown in Fig.~\ref{fig:comparison-approach}, a bi-encoder architecture~\cite{humeau2020polyencoders,reimers-gurevych-2019-sentence} encodes one embedding per document with the same model. Each sample in a batch has one positive pair. Negative pairs include all samples by different authors in a batch. InfoNCE loss~\cite{oord2018representation,wu2018unsupervised} with temperature scaling~\cite{chen2020simclr} is used to emphasize harder examples. This can be interpreted as a form of cross-entropy loss with temperature encouraging low similarity for negative pairs and high similarity for positive pairs, maximizing a lower bound on mutual information~\cite{oord2018representation} for same-class embeddings. It's equivalent to the NT-Xent loss~\cite{chen2020simclr} or Supervised Contrastive Loss~\cite{khosla2020supervised} with one positive pair. We use a SupCon implementation~\cite{musgrave2020pytorchmetric}. This is the loss for anchor $i$, positive $j^+$, batch negatives $j^- \in \mathcal{N}_i$, and temperature $\tau$.
\begin{equation}
    \mathcal{L}_{i} =
    - \log
    \frac{e^{s(\mathbf{z}_i, \mathbf{z}_{j^+}) / \tau}}
    {e^{s(\mathbf{z}_i, \mathbf{z}_{j^+}) / \tau}
    + \sum_{j^- \in \mathcal{N}_i}
    e^{s(\mathbf{z}_i, \mathbf{z}_{j^-}) / \tau}}
\end{equation}
Similarity $s(\mathbf{z}_i, \mathbf{z}_j)$ is computed using cosine similarity between the embeddings, allowing arbitrary pairs to be scored after encoding independently. The configuration used applies mean pooling over the final hidden states and doesn't use a projection head. For prediction, we select the cosine similarity threshold that maximizes F1 on the validation set.

\subsection {Experiments}

The validation set uses 10,000 same-author, different-fandom pairs and 10,000 different-author pairs with authors and fandoms removed from training data. Model selection, hyperparameters, and calibration use only the validation set.

A batch size and learning rate are selected for each approach by grid search on validation F1.  We use a form of \(\mu\)Transfer~\cite{yang2021tuning} where hyperparameters tuned on the smallest model, TinyBERT, are transferred to larger models. For transfer across depth, Complete(d)P~\cite{mlodozeniec2026completed} with \(\alpha=1\) justifies no adjustment for depth. In practice, we adjust only learning rate based on the change in width. This involves an adjustment factor of 0.4 going from width 312 (TinyBERT) to width 768 (for DistilBERT, BERT, and ModernBERT-base) and an adjustment factor of 0.3 going from width 312 (TinyBERT) to width 1024 (ModernBERT-large).

For simplicity not all hyperparameters are tuned~\cite{tuningplaybookgithub}, and we don't transfer weight decay, \(\beta_1\), \(\beta_2\), and \(\epsilon\) for AdamW but instead use recommendations or defaults from papers or code for each model. We don't apply weight decay to biases or layer normalization~\cite{devlin2019bert}. We use linear decay~\cite{bergsma2025straight} and 10\% warmup.

After fine-tuning for 10 epochs, both 256 and 512 context length results for contrastive models are shown in comparison to 512-length classification.

One model trained for 40 epochs, with Platt scaling and an abstention delta selected on the validation set, is used to report the full set of PAN21 test metrics.

\ifshowresultsdiscussion
\section {Results}

\subsection {Loss Function}
\begin{table}
\caption{TinyBERT contrastive, untuned batch size 256, untuned learning rate $\smash{2\times10^{-5}}$, PyTorch Metric Learning loss function defaults, 10 epochs.}
\label{tab:contrastive_loss_function_results}
\centering
\scriptsize
\setlength{\tabcolsep}{3pt}
\begin{tabular}{|p{3.2cm}|c|c|}
\hline
\textbf{Loss Function} & \textbf{Validation Accuracy} & \textbf{Validation F1} \\
\hline
InfoNCE & \textbf{0.753} & \textbf{0.774} \\
\hline
Circle & 0.732 & 0.764 \\
\hline
Multi-Similarity & 0.714 & 0.748 \\
\hline
Semi-Hard Contrastive & 0.704 & 0.734 \\
\hline
SoftTriple & 0.500 & 0.667 \\
\hline
Proxy Anchor & 0.500 & 0.667 \\
\hline
\end{tabular}
\end{table}
The loss functions listed in Table~\ref{tab:contrastive_loss_function_results} are surveyed by Musgrave et al.~\cite{musgrave2020metric}. InfoNCE loss has often been applied to larger datasets such as ImageNet~\cite{chen2020simclr} and Wikipedia~\cite{gao-2021-simcse}. It has been extended to emphasize harder instances~\cite{wu2018unsupervised}, include multiple positives~\cite{chen2020simclr,khosla2020supervised}, and function practically at larger batch sizes~\cite{gao-etal-2021-scaling}. It also has theoretical interpretations~\cite{oord2018representation,poole2019variational}. Pairwise contrastive loss has sometimes been applied to authorship verification~\cite{boenninghoff2019explainable,tyo2021siamese}. A pairwise contrastive loss, semi-hard contrastive~\cite{xuan2020easypositive}, performed worse than losses that can consider all negatives in a batch such as InfoNCE, Circle~\cite{sun2020circle}, and Multi-Similarity~\cite{wang2019multi}. SoftTriple~\cite{qian2019softtriple} and Proxy Anchor~\cite{kim2020proxyanchor}, based on learning proxy embeddings per class, fail in this setting with 238,815 author classes in the training set used.

\subsection {Model Comparison}
\begin{table}
\caption{Length 512, epoch 10, tuned models. Batch size 8, learning rate $\smash{4\times10^{-5}}$ for TinyBERT classification. Batch size 512, learning rate $\smash{4\times10^{-4}}$ for TinyBERT contrastive. Learning rate factor 0.3 for ModernBERT-large and 0.4 for other models.}
\label{tab:contrastive_vs_classification}
\centering
\scriptsize
\setlength{\tabcolsep}{4pt}
\begin{tabular}{|p{2.6cm}|p{1.7cm}|c|c|c|}
\hline
\textbf{Model} & \textbf{Objective} & \textbf{Test Accuracy} & \textbf{Test F1} & \textbf{Half-Length Test F1} \\
\hline
TinyBERT & Classification & 0.745 & 0.761 & \\
\hline
DistilBERT-cased & Classification & 0.684 & 0.725 & \\
\hline
BERT-base-cased & Classification & 0.787 & 0.797 & \\
\hline
ModernBERT-base & Classification & 0.653 & 0.706 & \\
\hline
ModernBERT-large & Classification & 0.703 & 0.742 & \\
\hline
TinyBERT & Contrastive & 0.812 & 0.812 & 0.789 \\
\hline
DistilBERT-cased & Contrastive & 0.877 & 0.874 & 0.822 \\
\hline
BERT-base-cased & Contrastive & 0.891 & 0.881 & 0.836 \\
\hline
ModernBERT-base & Contrastive & 0.900 & 0.898 & 0.841 \\
\hline
ModernBERT-large & Contrastive & \textbf{0.913} & \textbf{0.911} & 0.861 \\
\hline
\end{tabular}
\end{table}

As shown in Table~\ref{tab:contrastive_vs_classification}, contrastive learning outperformed even when limited to 256-token-length input context, while using less computation than 512-token-length classification. ModernBERT's classification performance could possibly be related to dropping the next sentence prediction task during pre-training.

\subsection {Mean Pooling and Data Augmentation}
\begin{table}
\caption{TinyBERT contrastive, batch size 1024, learning rate $\smash{4\times10^{-4}}$, temperature 0.01, 40 epochs. The projection head here uses a residual, GELU, and LayerNorm.}
\label{tab:contrastive_ablation}
\centering
\scriptsize
\setlength{\tabcolsep}{4pt}
\begin{tabular}{|p{7.5cm}|c|c|}
\hline
\textbf{Configuration} & \textbf{Val Acc} & \textbf{Val F1} \\
\hline
Two-Layer Projection Head + Mean First--Last Layer Pooling & 0.851 & 0.856 \\
\hline
Mean First--Last Layer Pooling only & 0.853 & 0.857 \\
\hline
Two-Layer Projection Head + Mean Pooling & 0.857 & 0.860 \\
\hline
Mean Pooling only & \textbf{0.861} & \textbf{0.863} \\
\hline
Mean Pooling only, no random text spans (6 epochs) & 0.713 & 0.740 \\
\hline
Mean Pooling only, no random text spans, 40 epochs & 0.682 & 0.727 \\
\hline
\end{tabular}
\end{table}

SimCSE~\cite{gao-2021-simcse} used mean first-last layer pooling; ablation in Table~\ref{tab:contrastive_ablation} shows better results here with the last layer only. This can be interpreted as forcing the model to use all the transformer layers and to use transformers directly for the representation with respect to every token. Longer duration training performs better with TinyBERT and uses a larger tuned batch size. The performance drop and overfitting without random text spans show the effectiveness of the data augmentation.

\subsection {Context Length and Batch Size}

As shown in Table~\ref{tab:contrastive_sequence_length}, a context length of 1024 tokens instead of 512 increases performance. It's left to future work to investigate whether longer context length effectively means that a larger batch size is supported by the additional data.

\begin{table}
\caption{ModernBERT-large Bi-Encoder, learning rate $\smash{1.2\times10^{-4}}$, temperature 0.01, 40 epochs.}
\label{tab:contrastive_sequence_length}
\centering
\scriptsize
\setlength{\tabcolsep}{4pt}
\begin{tabular}{|p{2.4cm}|p{1.7cm}|c|c|}
\hline
\textbf{Context Length} & \textbf{Batch Size} & \textbf{Test Accuracy} & \textbf{Test F1} \\
\hline
512 & 1024 & 0.926 & 0.925 \\
\hline
1024 & 1024 & 0.952 & 0.955 \\
\hline
2048 & 2048 & 0.976 & 0.975 \\
\hline
4096 & 4096 & \textbf{0.984} & \textbf{0.984} \\
\hline
\end{tabular}
\end{table}

\begin{table}
\caption{Comparison to prior work.}
\label{tab:comparison_to_prior_work}
\centering
\scriptsize
\setlength{\tabcolsep}{4pt}
\begin{tabular}{|p{4.2cm}|c|c|c|c|c|c|}
\hline
\textbf{Model} & \textbf{AUC} & \textbf{c@1} & \textbf{F1} & \textbf{F0.5u} & \textbf{Brier} & \textbf{Overall} \\
\hline
tyo21~\cite{tyo2021siamese} & 0.828 & 0.759 & 0.791 & 0.726 & 0.812 & 0.783 \\
\hline
weerasinghe21~\cite{weerasinghe2021feature} & 0.972 & 0.917 & 0.916 & 0.925 & 0.934 & 0.933 \\
\hline
peng21~\cite{peng2021encoding} & 0.917 & 0.917 & 0.917 &  0.920 & 0.917 & 0.918 \\
\hline
embarcaderoruiz21~\cite{embarcaderoruiz2022graph} & 0.970 & 0.931 & 0.934 & 0.915 & 0.931 & 0.936 \\
\hline
BigBird Cross-Encoder~\cite{nguyen-etal-2023-improving} & 0.990 & 0.946 & 0.944 & 0.962 & 0.956 & 0.960 \\
\hline
boenninghoff21~\cite{boenninghoff2021o2d2} & 0.987 & 0.950 & 0.952 & 0.938 & 0.945 & 0.955 \\
\hline
ModernBERT Bi-Encoder (Ours) & 0.995 & 0.982 & \textbf{0.984} & 0.960 & 0.972 & 0.979 \\
\hline
\end{tabular}
\end{table}

\subsection {Comparison to Prior Work}

As shown in Table~\ref{tab:comparison_to_prior_work}, Peng et al.'s concatenated input BERT model with classification achieves F1 of 0.917 that exceeds the F1 of contrastive BERT at 0.881, which can be attributed to using more than only 512 tokens per text. The results with BigBird Cross-Encoder and Boenninghoff et al.'s bidirectional LSTM, neither of which are constrained by a 512-token context length, also show the value of longer context on the fanfiction data that has up to 21,000 characters per document. The 4096 token context length ModernBERT Bi-Encoder achieves state-of-the-art performance. Based on the earlier comparison between classification and contrastive learning approaches, it is plausible that some of this can be attributed to the use of contrastive learning. The context length, model choice, training duration, and data augmentation also contribute to the result.

\section {Discussion}

Echoing earlier similarity-based feature engineering with many negatives~\cite{koppel2011authorship}, our results support similarity-based deep learning with many negatives. This outperformed the classification-based approach, but it is left to future work to consider performance on authorship attribution, style change detection, and other datasets. The results suggest that learning useful representations for authorial style can be done efficiently on each text independently. Computational advantages from delaying pairwise comparison until the cosine similarity metric used as part of the loss function appear to outweigh any benefits of attention across every pair. This method benefits from large batch sizes that have many negative instances. The contrastive learning objective for open-set authorship verification works as if solving many implicit authorship attribution proxy tasks. The interpretation here follows the view of InfoNCE~\cite{oord2018representation} as a categorical cross-entropy objective over one positive and many negative samples, where the model is trained to identify the positive example among the batch candidates.

\begin{credits}
\ifanonymous
\else
\subsubsection{\ackname} Special thanks to Eric Wang and Henry Luk for technical assistance and for comments on an earlier project report.
\subsubsection{\discintname}
The author has no competing interests.
\fi

\end{credits}

\else
\section {Results}

\section {Discussion}

\fi

%
%
%
%
\bibliographystyle{splncs04}
\bibliography{references}

\end{document}